\documentclass{article}

\usepackage{microtype}
\usepackage{pifont}
\usepackage{graphicx}
\usepackage{subfigure}
\usepackage{booktabs}
\usepackage{amsmath, amsthm, amssymb, amsfonts}
\usepackage{placeins}
\usepackage{stfloats}
\newtheorem{theorem}{Theorem}
\newtheorem{proposition}{Proposition}
\newtheorem*{theorem*}{Theorem}

\usepackage{siunitx}
\usepackage{chngcntr}
\usepackage{apptools}
\usepackage{caption}
\usepackage{enumitem}
\AtAppendix{\counterwithin{theorem}{section}}

\usepackage{hyperref}

\newcommand{\cmark}{\ding{51}}%
\newcommand{\xmark}{\ding{55}}%

\DeclareMathOperator*{\argmin}{arg\,min}
\DeclareMathOperator*{\argmax}{arg\,max}

\def\eg{{\em e.g.}}
\def\ie{{\em i.e.}}

\usepackage[accepted]{icml2019}

\icmltitlerunning{Similarity of Neural Network Representations Revisited}

\begin{document}

\twocolumn[
\icmltitle{Similarity of Neural Network Representations Revisited}

\begin{icmlauthorlist}
\icmlauthor{Simon Kornblith}{to}
\icmlauthor{Mohammad Norouzi}{to}
\icmlauthor{Honglak Lee}{to}
\icmlauthor{Geoffrey Hinton}{to}
\end{icmlauthorlist}

\icmlaffiliation{to}{Google Brain}

\icmlcorrespondingauthor{Simon Kornblith}{skornblith@google.com}

\icmlkeywords{Machine Learning, ICML}

\vskip 0.24in
]

\printAffiliationsAndNotice{}  %

\begin{abstract}
Recent work has sought to understand the behavior of neural networks by comparing representations between layers and between different trained models. We examine methods for comparing neural network representations based on canonical correlation analysis (CCA). We show that CCA belongs to a family of statistics for measuring multivariate similarity, but that neither CCA nor any other statistic that is invariant to invertible linear transformation can measure meaningful similarities between representations of higher dimension than the number of data points. We introduce a similarity index that measures the relationship between representational similarity matrices and does not suffer from this limitation. This similarity index is equivalent to centered kernel alignment (CKA) and is also closely connected to CCA. Unlike CCA, CKA can reliably identify correspondences between representations in networks trained from different initializations.
\end{abstract}

\vspace*{-.5cm}
\section{Introduction}
\vspace*{-.1cm}

Across a wide range of machine learning tasks, deep neural networks
enable learning powerful feature representations automatically from data.
Despite impressive empirical advances of deep neural networks in solving various tasks, the problem of understanding and characterizing the neural network representations learned from data remains relatively under-explored. 
Previous work (\eg~ \citet{advani2017high,amari2018dynamics,saxe2013exact}) has made progress in understanding the theoretical dynamics of the neural network training process. These studies are insightful, but fundamentally limited, because they ignore the complex interaction
between the training dynamics and structured data.
A window into the network's representation can provide more information about the interaction between machine learning algorithms and data than %
the value of the loss function alone. 

This paper investigates the problem of measuring similarities between deep neural network representations. 
An effective method for measuring representational similarity could help %
answer many interesting questions,
including: (1)~Do deep neural networks with the same architecture trained from different random initializations learn similar representations?
(2)~Can we establish correspondences between layers of different network architectures?
(3)~How similar are the representations learned using the same network architecture from different datasets?

We build upon previous studies investigating similarity between the representations of neural networks \cite{laakso2000content,pmlr-v44-li15convergent,svcca,morcos2018insights,wang2018}.
We are also inspired by the extensive neuroscience literature that uses representational similarity analysis \cite{kriegeskorte2008representational,edelman1998representation} to compare representations across brain areas \cite{haxby2001distributed,freiwald2010functional}, individuals \cite{connolly2012representation}, species \cite{kriegeskorte2008matching}, and behaviors \cite{elsayed2016reorganization}, as well as between brains and neural networks \cite{yamins2014performance,khaligh2014deep,sussillo2015neural}.

Our key contributions are summarized as follows:
\begin{itemize}[topsep=2pt, partopsep=0pt, leftmargin=15pt, parsep=0pt, itemsep=8pt]
    \item We discuss the invariance properties of similarity indexes and their implications for measuring similarity of neural network representations.
    \item We motivate and introduce {\em centered kernel alignment (CKA)} as a similarity index and analyze the relationship between CKA, linear regression, canonical correlation analysis (CCA), and related methods \cite{svcca,morcos2018insights}.
    \item We show that CKA is able to determine the correspondence between the hidden layers of neural networks trained from different random initializations and with different widths, scenarios where previously proposed similarity indexes fail.
    \item We verify that wider networks learn more similar representations, and show that the similarity of early layers saturates at fewer channels than later layers. We demonstrate that early layers, but not later layers, learn similar representations on different datasets.
\end{itemize}

\subsection*{Problem Statement}
Let $X \in \mathbb{R}^{n\times p_1}$ denote a matrix of activations of $p_1$ neurons for $n$ examples,
and $Y \in \mathbb{R}^{n\times p_2}$ denote a matrix of activations of $p_2$ neurons for the same $n$ examples.
We assume that these matrices have been preprocessed to center the columns.
Without loss of generality we assume that $p_1 \le p_2$.
We are concerned with the design and analysis of a scalar {\em similarity index} $s(X, Y)$ that can be used to compare representations within and across neural networks,
in order to help visualize and understand the effect of different factors of variation in deep learning.

\section{What Should Similarity Be Invariant To?}

This section discusses the invariance properties of similarity indexes and their implications for measuring similarity of neural network representations.
We argue that both intuitive notions of similarity and the dynamics of neural network training call for a similarity index that is invariant to orthogonal transformation and isotropic scaling, but not invertible linear transformation.

\subsection{Invariance to Invertible Linear Transformation}
\label{sec:linear_transform}
A similarity index is invariant to invertible linear transformation if $s(X, Y) = s(XA, YB)$ for any full rank $A$ and $B$.
If activations $X$ are followed by a fully-connected layer $f(X) = \sigma(XW + \beta)$, then transforming
the activations by a full rank matrix $A$ as $X' = XA$ and transforming the weights by the inverse $A^{-1}$ as $W' = A^{-1}W$ preserves the output of $f(X)$. 
This transformation does not appear to change how the network operates, so intuitively, one might prefer a similarity index that is invariant to invertible linear transformation, as argued by \citet{svcca}.

However, a limitation of invariance to invertible linear transformation is that any invariant similarity index gives the same result for any representation of width greater than or equal to the dataset size, \ie~$p_2 \geq n$.
We provide a simple proof in Appendix \ref{theorem_1_proof}.

\begin{theorem}
\label{invariance_theorem}
Let $X$ and $Y$ be $n \times p$ matrices. Suppose $s$ is invariant to invertible linear transformation in the first argument, \ie~$s(X, Z) = s(XA, Z)$ for arbitrary $Z$ and any $A$ with $\text{rank}(A) = p$. If $\text{rank}(X) = \text{rank}(Y) = n$, then $s(X, Z) = s(Y, Z)$.
\end{theorem}

There is thus a practical problem with invariance to invertible linear transformation: Some neural networks, especially convolutional networks, have more neurons in some layers than there are examples the training dataset \cite{springenberg2014striving,lee2018deep,wideresnet}. It is somewhat unnatural that a similarity index could require more examples than were used for training. %

A deeper issue is that neural network \textit{training} is not invariant to arbitrary invertible linear transformation of inputs or activations. Even in the linear case, gradient descent converges first along the eigenvectors corresponding to the largest eigenvalues of the input covariance matrix \cite{lecun1991second}, and in cases of overparameterization or early stopping, the solution reached depends on the scale of the input. Similar results hold for gradient descent training of neural networks in the infinite width limit \cite{jacot2018neural}. The sensitivity of neural networks training to linear transformation is further demonstrated by the popularity of batch normalization \cite{ioffe2015batch}.

\begin{figure}
    \centering
    \includegraphics[width=3.25in]{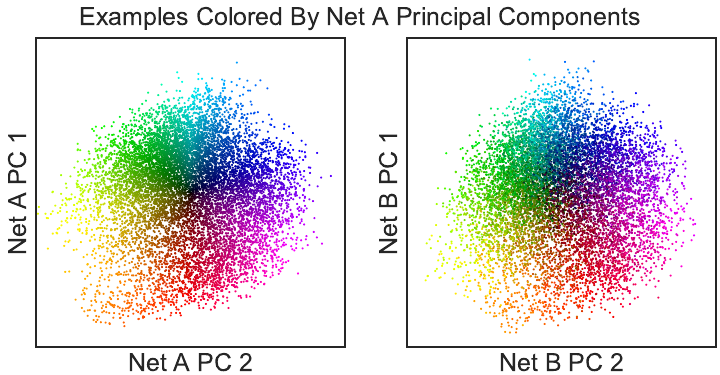}
    \vskip -0.05in
    \caption{First principal components of representations of networks trained from different random initializations are similar. 
    Each example from the CIFAR-10 test set is shown as a dot colored according to the value of the first two principal components of an intermediate layer of one network (left) and plotted on the first two principal components of the same layer of an architecturally identical network trained from a different initialization (right).}
    \label{fig:pcs}
    \vskip -0.1in
\end{figure}

Invariance to invertible linear transformation implies that the scale of directions in activation space is irrelevant. Empirically, however, scale information is both consistent across networks and useful across tasks. Neural networks trained from different random initializations develop representations with similar large principal components, as shown in Figure~\ref{fig:pcs}. Consequently, Euclidean distances between examples, which depend primarily upon large principal components, are similar across networks. These distances are \textit{meaningful}, as demonstrated by the success of perceptual loss and style transfer \cite{gatys2016image,johnson2016perceptual,dumoulin2017learned}. %
A similarity index that is invariant to invertible linear transformation ignores this aspect of the representation, and assigns the same score to networks that match only in large principal components
or
networks that match only in small principal components.

\subsection{Invariance to Orthogonal Transformation}
\label{orthogonal_invariance_section}

Rather than requiring invariance to any invertible linear transformation, one could require a weaker condition; invariance to orthogonal transformation, \ie~$s(X, Y) = s(XU, YV)$ for full-rank orthonormal matrices $U$ and $V$ such that 
$U^\text{T}U = I$ and $V^\text{T}V = I$. 

Indexes invariant to orthogonal transformations do not share the limitations of indexes invariant to invertible linear transformation. When $p_2 > n$, indexes invariant to orthogonal transformation remain well-defined. Moreover, orthogonal transformations preserve scalar products and Euclidean distances between examples.

Invariance to orthogonal transformation seems desirable for neural networks trained by gradient descent. Invariance to orthogonal transformation implies invariance to permutation, which is needed to accommodate symmetries of neural networks \cite{chen1993geometry,orhan2018skip}. In the linear case, orthogonal transformation of the input does not affect the dynamics of gradient descent training \cite{lecun1991second}, and for neural networks initialized with rotationally symmetric weight distributions, \eg~i.i.d. Gaussian weight initialization, training with fixed orthogonal transformations of activations yields the same distribution of training trajectories as untransformed activations, whereas an arbitrary linear transformation would not.

Given a similarity index $s(\cdot, \cdot)$ that is invariant to orthogonal transformation, one can construct a similarity index $s'(\cdot, \cdot)$ that is invariant to any invertible linear transformation
by first orthonormalizing the columns of $X$ and $Y$, and then applying $s(\cdot, \cdot)$. Given thin QR decompositions $X = Q_AR_A$ and $Y = Q_BR_B$ %
one can construct a similarity index $s'(X,Y) = s(Q_X, Q_Y)$, where $s'(\cdot, \cdot)$ is invariant
to invertible linear transformation because orthonormal bases with the same span are related to each other by orthonormal transformation (see Appendix~\ref{orthogonality_invariance_theorem}).

\subsection{Invariance to Isotropic Scaling}

We expect similarity indexes to be invariant to isotropic scaling, \ie~$s(X, Y) = s(\alpha X, \beta Y)$ for any $\alpha, \beta \in \mathbb{R}^{+}$. That said, a similarity index that is invariant to both orthogonal transformation and non-isotropic scaling,
\ie~rescaling of individual features, is invariant to any invertible linear transformation. This follows from the existence of the singular value decomposition of the transformation matrix.
Generally, we are interested in similarity indexes that are invariant to isotropic but not necessarily non-isotropic scaling.

\section{Comparing Similarity Structures}

Our key insight is that instead of comparing multivariate features of an example in the two representations (\eg~via regression),
one can first measure the similarity between every pair of \textit{examples} in each representation separately,
and then compare the similarity structures. In neuroscience, such matrices representing the similarities between examples are called representational similarity matrices \cite{kriegeskorte2008representational}.
We show below that, if we use an inner product to measure similarity, the similarity between representational similarity matrices reduces to another intuitive notion of pairwise feature similarity.

\paragraph{Dot Product-Based Similarity.}
A simple formula relates dot products between examples to dot products between features:
\begin{align}\label{linear_ss}
\langle\text{vec}(XX^\text{T}),\text{vec}(YY^\text{T})\rangle = \text{tr}(XX^\text{T}YY^\text{T}) = ||Y^\text{T}X||_\text{F}^2.
\end{align}
The elements of $XX^\text{T}$ and $YY^\text{T}$ are dot products between the representations of the $i^\text{th}$ and $j^\text{th}$ examples, and indicate the similarity between these examples according to the respective networks. The left-hand side of \eqref{linear_ss} thus measures the similarity between the inter-example similarity structures. The right-hand side yields the same result by measuring the similarity between \textit{features} from $X$ and $Y$, by summing the squared dot products between every pair.

\paragraph{Hilbert-Schmidt Independence Criterion.}
\label{sec:hsic}
Equation \ref{linear_ss} implies that, for centered $X$ and $Y$:
\begin{align}\label{cov_linear_ss} \frac{1}{(n-1)^2}\text{tr}(XX^\text{T}YY^\text{T}) &=  ||\text{cov}(X^\text{T}, Y^\text{T})||_\text{F}^2.
\end{align}

The Hilbert-Schmidt Independence Criterion \cite{gretton2005measuring} generalizes Equations \ref{linear_ss} and \ref{cov_linear_ss} to inner products from reproducing kernel Hilbert spaces, where the squared Frobenius norm of the cross-covariance matrix becomes the squared Hilbert-Schmidt norm of the cross-covariance operator. Let $K_{ij} = k(\mathbf{x}_i, \mathbf{x}_j)$ and $L_{ij} = l(\mathbf{y}_i, \mathbf{y}_j)$ where $k$ and $l$ are two kernels. The empirical estimator of HSIC is:
\begin{align}
\text{HSIC}(K, L) = \frac{1}{(n-1)^2}\text{tr}(KHLH),
\end{align}
where $H$ is the centering matrix $H_n = I_n - \frac{1}{n} \mathbf{1}\mathbf{1}^\text{T}$. For linear kernels $k(\mathbf{x}, \mathbf{y}) = l(\mathbf{x}, \mathbf{y}) = \mathbf{x}^\text{T}\mathbf{y}$, HSIC yields \eqref{cov_linear_ss}.

\citet{gretton2005measuring} originally proposed HSIC as a test statistic for determining whether two sets of variables are independent. They prove that the empirical estimator converges to the population value at a rate of $1/\sqrt{n}$, and \citet{song2007supervised} provide an unbiased estimator.
When $k$ and $l$ are universal kernels, HSIC = 0 implies independence, but HSIC is not an estimator of mutual information. HSIC is equivalent to maximum mean discrepancy between the joint distribution and the product of the marginal distributions, and HSIC with a specific kernel family is equivalent to distance covariance \cite{sejdinovic2013equivalence}.

\paragraph{Centered Kernel Alignment.}
HSIC is not invariant to isotropic scaling, but it can be made invariant through normalization. This normalized index is known as centered kernel alignment \cite{cortes2012algorithms,cristianini2002kernel}:
\begin{align}
    \text{CKA}(K, L) = \frac{\text{HSIC}(K, L)}{\sqrt{\text{HSIC}(K, K)\text{HSIC}(L, L)}}.
\end{align}

For a linear kernel, CKA is equivalent to the RV coefficient \cite{robert1976unifying} and to Tucker's congruence coefficient \cite{tucker1951method,lorenzo2006tucker}.

\paragraph{Kernel Selection.}
Below, we report results of CKA with a linear kernel and the RBF kernel $k(\textbf{x}_i, \textbf{x}_j) = \exp(-||\textbf{x}_i - \textbf{x}_j||_2^2/(2\sigma^2))$. For the RBF kernel, there are several possible strategies for selecting the bandwidth $\sigma$, which controls the extent to which similarity of small distances is emphasized over large distances. We set $\sigma$ as a fraction of the median distance between examples. In practice, we find that RBF and linear kernels give similar results across most experiments, so we use linear CKA unless otherwise specified. Our framework extends to any valid kernel, including kernels equivalent to neural networks \cite{lee2018deep,jacot2018neural,garriga2018deep,novak2019bayesian}.

\section{Related Similarity Indexes}

\begin{table*}
    \centering
    \setlength{\tabcolsep}{0.5em}
    \begin{tabular}{|l|l|c|c|c|}
        \cline{3-5}
        \multicolumn{1}{c}{} & \multicolumn{1}{c|}{} & \multicolumn{3}{c|}{\centering \bf Invariant to} \\
        \hline & & \bf Invertible Linear & \bf Orthogonal & \bf Isotropic \\
        \multicolumn{1}{|c|}{\bf Similarity Index} & \multicolumn{1}{c|}{\bf Formula} & \bf Transform & \bf Transform & \bf Scaling\\ \hline
         Linear Reg. ($R^2_\text{LR}$) & $||Q_Y^\text{T}X||_\text{F}^2/||X||_\text{F}^2$ & $Y$ only & \cmark & \cmark\\
         CCA ($R^2_\text{CCA}$) & $||Q_Y^\text{T}Q_X||_\text{F}^2/p_1$ & \cmark & \cmark & \cmark\\
         CCA ($\bar{\rho}_\text{CCA}$) & $||Q_Y^\text{T}Q_X||_*/p_1$ & \cmark & \cmark & \cmark\\
         SVCCA ($R^2_\text{SVCCA}$) & $||(U_Y T_Y)^\text{T}U_X T_X||_\text{F}^2/\text{min}(||T_X||^2_\text{F},||T_Y||^2_\text{F})$ & If same subspace kept & \cmark & \cmark\\
         SVCCA ($\bar{\rho}_\text{SVCCA}$) & $||(U_YT_Y)^\text{T}U_XT_X ||_*/\text{min}(||T_X||^2_\text{F},||T_Y||^2_\text{F})$ & If same subspace kept & \cmark & \cmark\\
         PWCCA & $\sum_{i=1}^{p_1} \alpha_i \rho_i/||\mathbf{\alpha}||_1$, $
\alpha_i = \sum_{j} |\langle \mathbf{h}_i, \mathbf{x}_j \rangle|$ & \xmark & \xmark & \cmark\\
         Linear HSIC & $||Y^\text{T}X||_\text{F}^2/(n-1)^2$ & \xmark & \cmark & \xmark\\
         Linear CKA & $||Y^\text{T}X||_\text{F}^2/(||X^\text{T}X||_\text{F}||Y^\text{T}Y||_\text{F})$ & \xmark & \cmark & \cmark\\
         RBF CKA & $\text{tr}(KHLH)/\sqrt{\text{tr}(KHKH)\text{tr}(LHLH)}$ & \xmark & \cmark & \cmark$^*$\\
         \hline
    \end{tabular}
    \caption{Summary of similarity methods investigated. $Q_X$ and $Q_Y$ are orthonormal bases for the columns of $X$ and $Y$. $U_X$ and $U_Y$ are the left-singular vectors of $X$ and $Y$ sorted in descending order according to the corresponding singular vectors. $||\cdot||_*$~denotes the nuclear norm. $T_X$ and $T_Y$ are truncated identity matrices that select left-singular vectors such that the cumulative variance explained reaches some threshold. For RBF CKA, $K$ and $L$ are kernel matrices constructed by evaluating the RBF kernel between the examples as in Section~\ref{sec:hsic}, and $H$ is the centering matrix $H_n = I_n - \frac{1}{n} \mathbf{1}\mathbf{1}^\text{T}$. See Appendix~\ref{cca_and_linear_regression} for more detail about each technique.\\\
    $^*$Invariance of RBF CKA to isotropic scaling depends on the procedure used to select the RBF kernel bandwidth parameter. In our experiments, we selected the bandwidth as a fraction of the median distance, which ensures that the similarity index is invariant to isotropic scaling.}
    \label{tab:summary}
\end{table*}

In this section, we briefly review linear regression, canonical correlation, and other related methods in the context of measuring similarity between neural network representations. We let $Q_X$ and $Q_Y$ represent any orthonormal bases for the columns of $X$ and $Y$, \ie~$Q_X = X(X^\text{T}X)^{-1/2}$, $Q_Y = Y(Y^\text{T}Y)^{-1/2}$ or orthogonal transformations thereof. Table~\ref{tab:summary} summarizes the formulae and invariance properties of the indexes used in experiments. For a comprehensive general review of linear indexes for measuring multivariate similarity, see \citet{ramsay1984matrix}.

\paragraph{Linear Regression.}
A simple way to relate neural network representations is via linear regression. One can fit every feature in $X$ as a linear combination of features from $Y$. A suitable summary statistic is the total fraction of variance explained by the fit:
\begin{equation}
R_\text{LR}^2 = 1 - \frac{\min_B ||X - YB||_\text{F}^2}{||X||_\text{F}^2} = \frac{||Q_Y^\text{T}X||_\text{F}^2}{||X||_\text{F}^2}. \label{eq:r2lr}
\end{equation}
We are unaware of any application of linear regression to measuring similarity of neural network representations, although \citet{romero2014fitnets} used a least squares loss between activations of two networks to encourage thin and deep ``student" networks to learn functions similar to wide and shallow ``teacher" networks.

\paragraph{Canonical Correlation Analysis (CCA).}

Canonical correlation finds bases for two matrices such that, when the original matrices are projected onto these bases, the correlation is maximized. For $1 \leq i \leq p_1$, the $i$\textsuperscript{th} canonical correlation coefficient $\rho_i$ is given by:
\begin{equation}
\begin{aligned}
\rho_i = \max_{\mathbf{w}_X^i, \mathbf{w}_Y^i}&  \text{corr}(X\mathbf{w}_X^i, Y\mathbf{w}_Y^i)\\
\mathrm{subject~to} &\ ~~ \forall_{j< i}~~ X\mathbf{w}_X^i\perp X\mathbf{w}_X^j\\
&\ ~~ \forall_{j< i}~~ Y\mathbf{w}_Y^i\perp Y\mathbf{w}_Y^j .
\end{aligned}
\label{eq:eq1}
\end{equation}
The vectors $\mathbf{w}_X^i \in \mathbb{R}^{p_1}$ and $\mathbf{w}_Y^i \in \mathbb{R}^{p_2}$ that maximize $\rho_i$ are the canonical weights, which transform the original data into canonical variables $X\mathbf{w}_X^i$ and $Y\mathbf{w}_Y^i$. The constraints in \eqref{eq:eq1} enforce orthogonality of the canonical variables.

For the purpose of this work, we consider two summary statistics of the goodness of fit of CCA:
\begin{align}
    R^2_\text{CCA} &= \frac{\sum_{i=1}^{p_1} \rho_i^2}{p_1} = \frac{||Q_Y^\text{T}Q_X||_\text{F}^2}{p_1}\label{eq:r2cca}\\
    \bar \rho_\text{CCA} &= \frac{\sum_{i=1}^{p_1} \rho_i}{p_1} = \frac{||Q_Y^\text{T}Q_X||_*}{p_1} ,
\end{align}
where $||\cdot||_*$~denotes the nuclear norm. The mean squared CCA correlation $R^2_\text{CCA}$ is also known as {\em Yanai's GCD measure} \cite{ramsay1984matrix}, and several statistical packages report the sum of the squared canonical correlations $p_1 R_\text{CCA}^2 = \sum_{i=1}^{p_1} \rho_i^2$ under the name {\em Pillai's trace} \cite{sas_cca,stata_cca}.
The mean CCA correlation $\bar \rho_\text{CCA}$ was previously used to measure similarity between neural network representations in \citet{svcca}.

\paragraph{SVCCA.}

\label{sec:svcca}

CCA is sensitive to perturbation when the condition number of $X$ or $Y$ is large \cite{golub1995canonical}. To improve robustness, {\em singular vector CCA} (SVCCA) performs CCA on truncated singular value decompositions of $X$ and $Y$ \cite{svcca,mroueh2015asymmetrically,kuss2003geometry}.
As formulated in \citet{svcca}, SVCCA keeps enough principal components of the input matrices to explain a fixed proportion of the variance, and drops remaining components. Thus, it is invariant to invertible linear transformation only if the retained subspace does not change.

\paragraph{Projection-Weighted CCA.}

\citet{morcos2018insights} propose a different strategy to reduce the sensitivity of CCA to perturbation, which they term ``projection-weighted canonical correlation" (PWCCA):
\begin{align}
\rho_\text{PW} &= \frac{\sum_{i=1}^c \alpha_i \rho_i}{\sum_{i=1} \alpha_i} & 
\alpha_i &=\sum_{j} |\langle \mathbf{h}_i, \mathbf{x}_j \rangle| ,
\end{align}
 where $\mathbf{x}_j$ is the $j^\text{th}$ column of $X$, and $\mathbf{h}_i = X\mathbf{w}_X^i$ is the vector of canonical variables formed by projecting $X$ to the $i^\text{th}$ canonical coordinate frame. As we show in Appendix \ref{pwcca}, PWCCA is closely related to linear regression, since:
\begin{align}
R_\text{LR}^2 &= \frac{\sum_{i=1}^c \alpha_i' \rho_i^2}{\sum_{i=1} \alpha_i'} & 
\alpha_i' &=\sum_{j} \langle \mathbf{h}_i, \mathbf{x}_j \rangle^2 .
\end{align}

\paragraph{Neuron Alignment Procedures.}
Other work has studied alignment between individual neurons, rather than alignment between subspaces. \citet{pmlr-v44-li15convergent} examined correlation between the neurons in different neural networks, and attempt to find a bipartite match or semi-match that maximizes the sum of the correlations between the neurons, and then to measure the average correlations. \citet{wang2018} proposed to search for subsets of neurons $\tilde X \subset X$ and $\tilde Y \subset Y$ such that, to within some tolerance, every neuron in $\tilde X$ can be represented by a linear combination of neurons from $\tilde Y$ and vice versa.
They found that the maximum matching subsets are very small for intermediate layers.

\paragraph{Mutual Information.}
Among non-linear measures, one candidate is mutual information, which is invariant not only to invertible linear transformation, but to any invertible transformation. \citet{pmlr-v44-li15convergent} previously used mutual information to measure neuronal alignment. In the context of comparing representations, we believe mutual information is not useful. Given any pair of representations produced by deterministic functions of the same input, mutual information between either and the input must be at least as large as mutual information between the representations. Moreover, in fully invertible neural networks \cite{dinh2016density,jacobsen2018revnet}, the mutual information between any two layers is equal to the entropy of the input.

\section{Linear CKA versus CCA and Regression}

Linear CKA is closely related to CCA and linear regression. If $X$ and $Y$ are centered, then $Q_X$ and $Q_Y$ are also centered, so:
\begin{align}
R^2_\text{CCA} = \text{CKA}(Q_XQ_X^\text{T}, Q_YQ_Y^\text{T}) \sqrt{\frac{p_2}{p_1}} .
\end{align}
When performing the linear regression fit of $X$ with design matrix $Y$, $R_\text{LR}^2 = ||Q_Y^\text{T}X||_F^2/||X||_F^2$, so:
\begin{align}
R^2_\text{LR} = \text{CKA}(XX^\text{T}, Q_YQ_Y^\text{T}) \frac{\sqrt{p_1} ||X^\text{T}X||_\text{F}}{||X||_\text{F}^2} .
\end{align}

When might we prefer linear CKA over CCA? One way to show the difference is to rewrite $X$ and $Y$ in terms of their singular value decompositions $X = U_X\Sigma_XV_X^\text{T}$,  $Y = U_Y\Sigma_YV_Y^\text{T}$. Let the $i^\text{th}$ eigenvector of $XX^\text{T}$ (left-singular vector of $X$) be indexed as $\textbf{u}_X^\text{i}$. Then $R^2_\text{CCA}$ is:
\begin{align}
    R^2_\text{CCA} = ||U_Y^\text{T}U_X||_\text{F}^2/p_1 = \sum_{i=1}^{p_1} \sum_{j=1}^{p_2} \langle \textbf{u}_X^i, \textbf{u}_Y^j\rangle^2 / p_1 .
    \label{eq:cca_eigs}
\end{align}
Let the $i^\text{th}$ eigenvalue of $XX^\text{T}$ (squared singular value of $X$) be indexed as $\lambda_X^i$. Linear CKA can be written as:
\begin{align}
    \text{CKA}(XX^\text{T}, YY^\text{T}) &= \frac{||Y^\text{T}X||_\text{F}^2}{||X^\text{T}X||_\text{F}||Y^\text{T}Y||_\text{F}} \nonumber\\
    \label{eq:kta_eigs}
    &= \frac{\sum_{i=1}^{p_1} \sum_{j=1}^{p_2} \lambda_X^i \lambda_Y^j \langle \textbf{u}_X^i, \textbf{u}_Y^j\rangle^2}{\sqrt{\sum_{i=1}^{p_1} (\lambda_X^i)^2}\sqrt{\sum_{j=1}^{p_2} (\lambda_Y^j)^2}} .
\end{align}
Linear CKA thus resembles CCA weighted by the eigenvalues of the corresponding eigenvectors, \ie~the amount of variance in $X$ or $Y$ that each explains. SVCCA \cite{svcca} and projection-weighted CCA \cite{morcos2018insights} were also motivated by the idea that eigenvectors that correspond to small eigenvalues are less important, but linear CKA incorporates this weighting symmetrically and can be computed without a matrix decomposition.

Comparison of \eqref{eq:cca_eigs} and \eqref{eq:kta_eigs} immediately suggests the possibility of alternative weightings of scalar products between eigenvectors. Indeed, as we show in Appendix~\ref{appendix:regularized_cca}, the similarity index induced by ``canonical ridge" regularized CCA \cite{vinod1976canonical}, when appropriately normalized, interpolates between $R^2_\text{CCA}$, linear regression, and linear CKA. %

\section{Results}

\subsection{A Sanity Check for Similarity Indexes}
\label{sec:sanity_check}
\begin{figure}[t]
    \centering
    \includegraphics[width=3.25in]{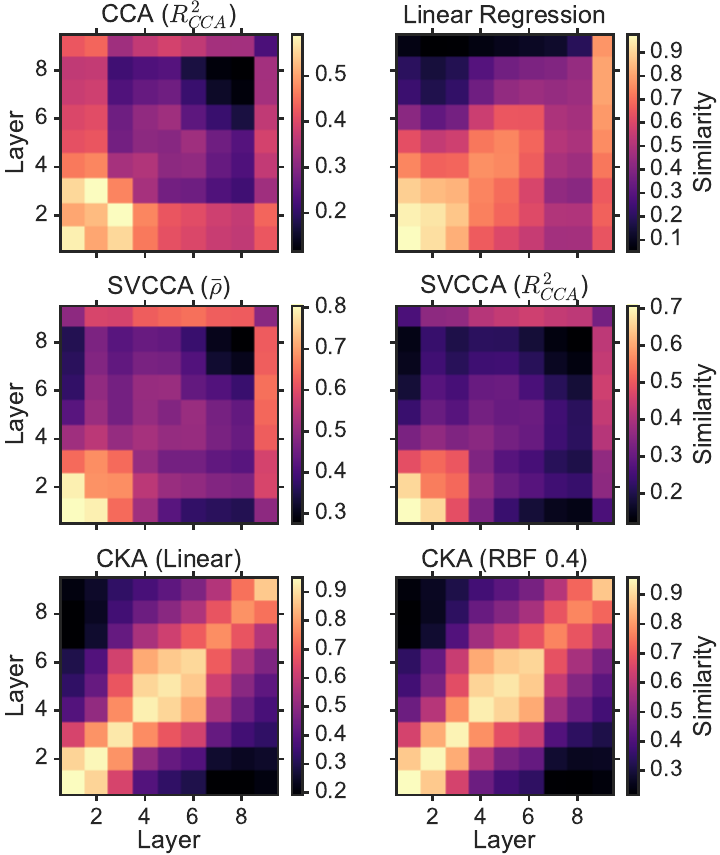}
    \vskip -0.05in
    \caption{CKA reveals consistent relationships between layers of CNNs trained with different random initializations, whereas CCA, linear regression, and SVCCA do not. For linear regression, which is asymmetric, we plot $R^2$ for the fit of the layer on the x-axis with the layer on the y-axis. Results are averaged over 10 networks on the CIFAR-10 training set. See Table \ref{tab:kta_eval} for a numerical summary.}
    \label{fig:comparison}
    \vskip -0.1in
\end{figure}

\begin{table}
\centering
\begin{small}
{
\begin{tabular}{lr}
    \toprule
    Index &  Accuracy\\
     \midrule
    CCA ($\bar \rho$) & 1.4\\
    CCA ($R^2_\text{CCA}$) & 10.6\\
    SVCCA ($\bar \rho$) & 9.9\\
    SVCCA ($R^2_\text{CCA}$) & 15.1\\
    PWCCA & 11.1\\
    Linear Reg. & 45.4\\
    Linear HSIC & 22.2\\
    CKA (Linear) & \textbf{99.3}\\
    CKA (RBF 0.2) & 80.6\\
    CKA (RBF 0.4) & \textbf{99.1}\\
    CKA (RBF 0.8) & \textbf{99.3}\\
    \bottomrule
\end{tabular}
}
\caption{Accuracy of identifying corresponding layers based on maximum similarity for 10 architecturally identical 10-layer CNNs trained from different initializations, with logits layers excluded. For SVCCA, we used a truncation threshold of 0.99 as recommended in \citet{svcca}. For asymmetric indexes (PWCCA and linear regression) we symmetrized the similarity as $S + S^\text{T}$. CKA RBF kernel parameters reflect the fraction of the median Euclidean distance used as $\sigma$. Results not significantly different from the best result are bold-faced ($p < 0.05$, jackknife z-test).}
\label{tab:kta_eval}
\end{small}
\vskip -0.1in
\end{table}

We propose a simple sanity check for similarity indexes: Given a pair of architecturally identical networks trained from different random initializations, for each layer in the first network, the most similar layer in the second network should be the architecturally corresponding layer. We train 10 networks and, for each layer of each network, we compute the accuracy with which we can find the corresponding layer in each of the other networks by maximum similarity. We then average the resulting accuracies.
We compare CKA with CCA, SVCCA, PWCCA, and linear regression. %

We first investigate a simple VGG-like convolutional network based on All-CNN-C \citep{springenberg2014striving}~(see Appendix \ref{architecture_details}) on CIFAR-10. Figure~\ref{fig:comparison} and Table~\ref{tab:kta_eval} show that CKA passes our sanity check, %
but other methods perform substantially worse. For SVCCA, we experimented with a range of truncation thresholds, but no threshold revealed the layer structure (Appendix~\ref{app:other_thresholds}); our results are consistent with those in Appendix E of \citet{svcca}.

We also investigate Transformer networks, where all layers are of equal width. In Appendix \ref{appendix:transformer}, we show similarity between the 12 sublayers of the encoders of Transformer models \citep{vaswani2017attention} trained from different random initializations. All similarity indexes achieve non-trivial accuracy and thus pass the sanity check, although RBF CKA and $R^2_\text{CCA}$ performed slightly better than other methods. However, we found that there are differences in feature scale between representations of feed-forward network and self-attention sublayers that CCA does not capture because it is invariant to non-isotropic scaling.

\subsection{Using CKA to Understand Network Architectures}

\begin{figure*}[t!]
    \centering
    \includegraphics[width=6.75in]{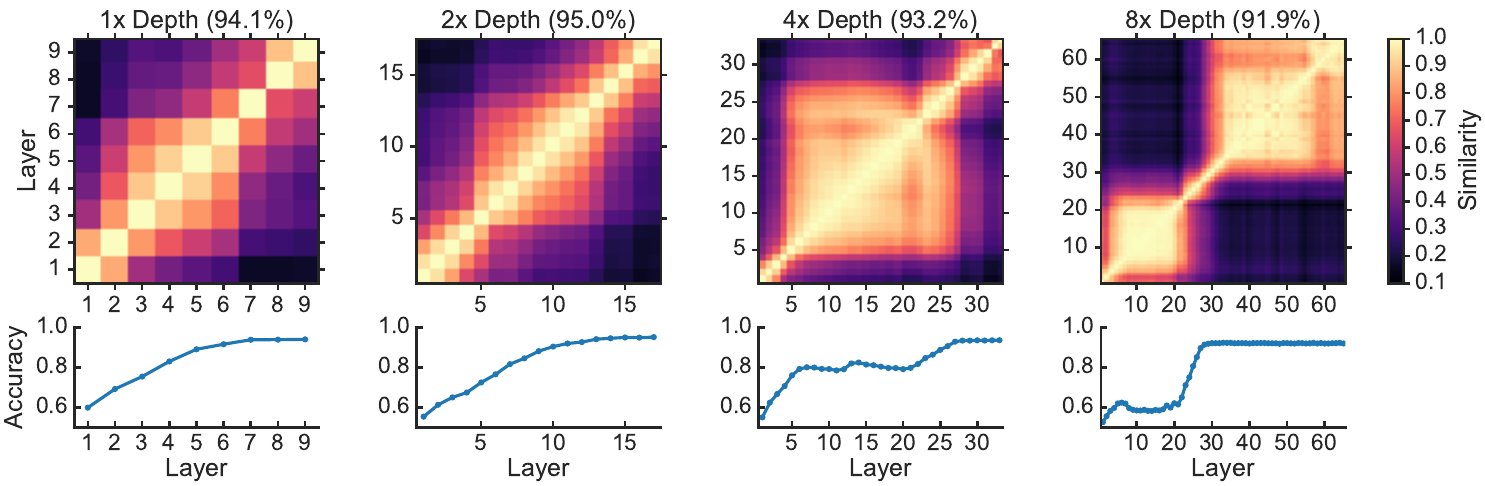}
    \vskip -0.1in
    \caption{CKA reveals when depth becomes pathological. \textbf{Top}: Linear CKA between layers of individual networks of different depths on the CIFAR-10 test set. Titles show accuracy of each network. Later layers of the 8x depth network are similar to the last layer. \textbf{Bottom}: Accuracy of a logistic regression classifier trained on layers of the same networks is consistent with CKA.}
    \label{fig:different_depths}
    \vskip -0.05in
\end{figure*}

\begin{figure}[t!]
    \centering
    \includegraphics[width=3.25in]{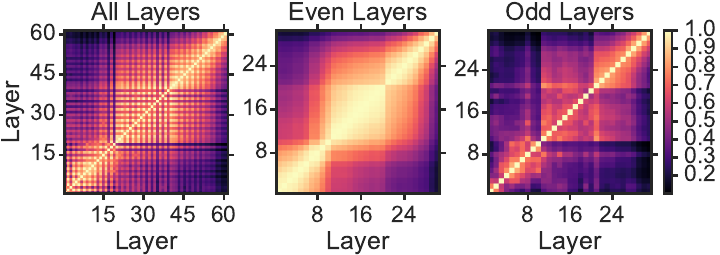}
    \vskip -0.05in
    \caption{Linear CKA between layers of a ResNet-62 model on the CIFAR-10 test set. The grid pattern in the left panel arises from the architecture. Right panels show similarity separately for even layer (post-residual) and odd layer (block interior) activations. Layers in the same block group (\ie~at the same feature map scale) are more similar than layers in different block groups.}
    \label{fig:resnets}
    \vskip -0.15in
\end{figure}

CKA can reveal pathology in neural networks representations. In Figure~\ref{fig:different_depths}, we show CKA between layers of individual CNNs with different depths, where layers are repeated 2, 4, or 8 times. Doubling depth improved accuracy, but greater multipliers hurt accuracy. At 8x depth, CKA indicates that representations of more than half of the network are very similar to the last layer. We validated that these later layers do not refine the representation by training an $\ell^2$-regularized logistic regression classifier on each layer of the network. Classification accuracy in shallower architectures progressively improves with depth, but for the 8x deeper network, accuracy plateaus less than halfway through the network. When applied to ResNets \cite{he2016deep}, CKA reveals no pathology (Figure~\ref{fig:resnets}). We instead observe a grid pattern that originates from the architecture: Post-residual activations are similar to other post-residual activations, but activations within blocks are not.

\begin{figure}[t!]
    \centering
    \includegraphics[width=3.25in]{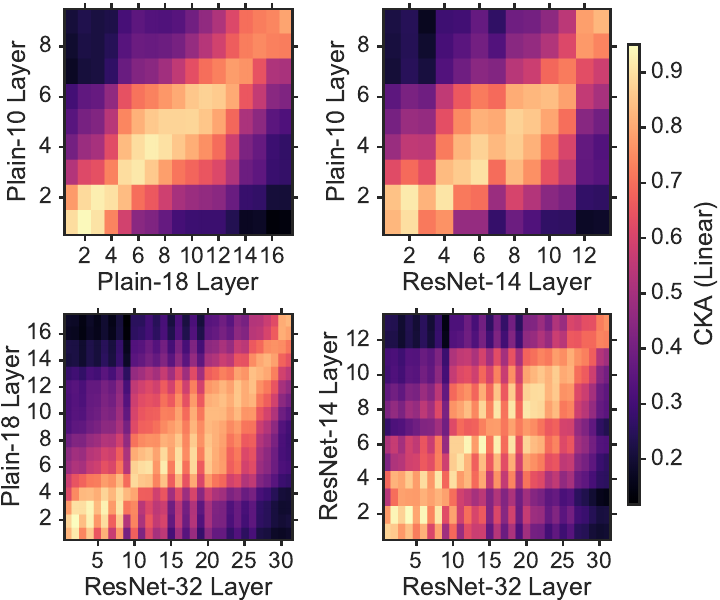}
    \vskip -0.05in
    \caption{Linear CKA between layers of networks with different architectures on the CIFAR-10 test set.}
    \label{fig:cross_arch}
\end{figure}

CKA is equally effective at revealing relationships between layers of different architectures. Figure \ref{fig:cross_arch} shows the relationship between different layers of networks with and without residual connections. CKA indicates that, as networks are made deeper, the new layers are effectively inserted in between the old layers. Other similarity indexes fail to reveal meaningful relationships between different architectures, as we show in Appendix~\ref{app:tiny_10_v_resnet}.

\begin{figure}[t!]
    \centering
    \vskip 0.1in
    \includegraphics[width=3.25in]{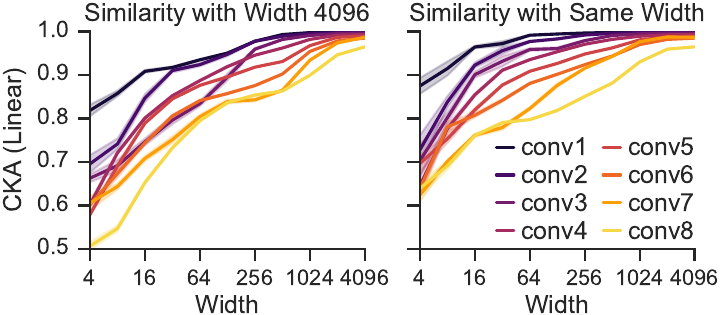}
    \vskip -0.05in
    \caption{Layers become more similar to each other and to wide networks as width increases, but similarity of earlier layers saturates first. \textbf{Left}: Similarity of networks with the widest network we trained. \textbf{Middle}: Similarity of networks with other networks of the same width trained from random initialization. All CKA values are computed between 10 networks on the CIFAR-10 test set; shaded regions reflect jackknife standard error.}
    \label{fig:width_experiments}
    \vskip -0.25in
\end{figure}

In Figure \ref{fig:width_experiments}, we show CKA between networks with different layer widths. Like \citet{morcos2018insights}, we find that increasing layer width leads to more similar representations between networks. As width increases, CKA approaches 1; CKA of earlier layers saturates faster than later layers. Networks are generally more similar to other networks of the same width than they are to the widest network we trained.

\subsection{Similar Representations Across Datasets}

\begin{figure}[t]
    \centering
    \includegraphics[width=3.25in]{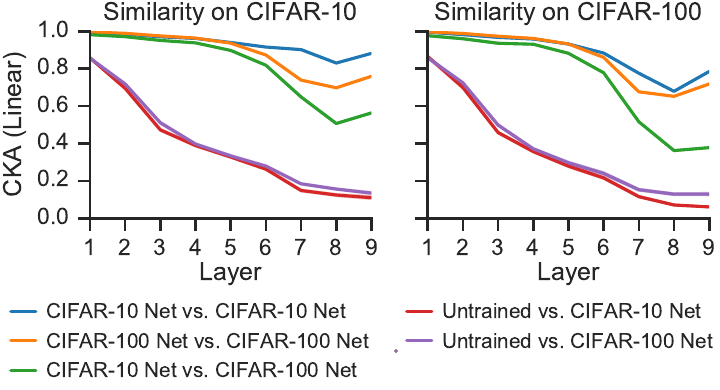}
    \vskip -0.05in
    \caption{CKA shows that models trained on different datasets (CIFAR-10 and CIFAR-100) develop similar representations, and these representations differ from untrained models. The left panel shows similarity between the same layer of different models on the CIFAR-10 test set, while the right panel shows similarity computed on CIFAR-100 test set. CKA is averaged over 10 models of each type (45 pairs).}
    \label{fig:cross_dataset}
    \vskip -0.15in
\end{figure}

CKA can also be used to compare networks trained on different datasets. In Figure~\ref{fig:cross_dataset}, we show that models trained on CIFAR-10 and CIFAR-100 develop similar representations in their early layers. These representations require training; similarity with untrained networks is much lower. We further explore similarity between layers of untrained networks in Appendix~\ref{appendix:cka_init}.

\subsection{Analysis of the Shared Subspace}
\begin{figure}[t]
    \centering
    \includegraphics[width=3.25in]{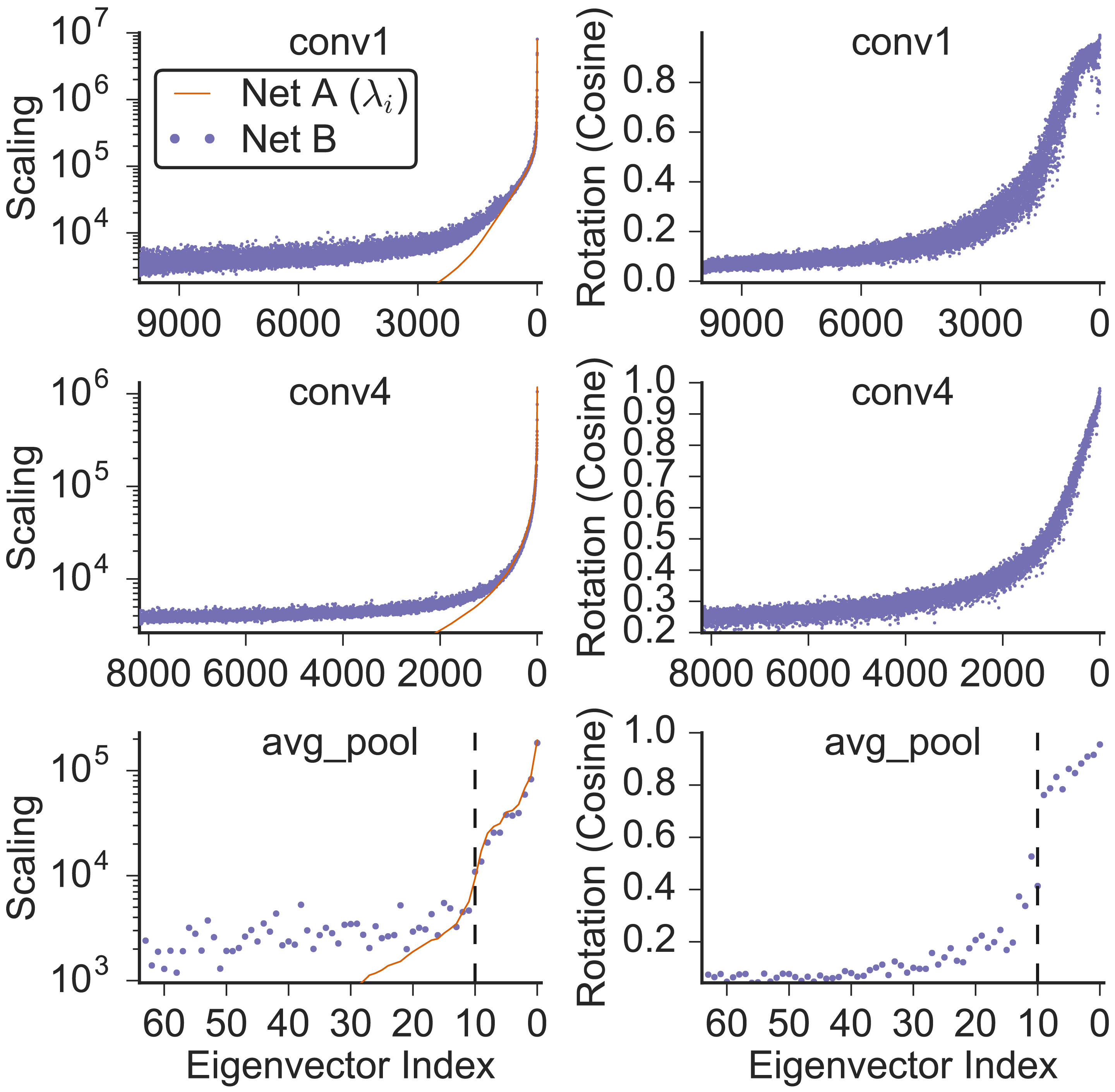}
    \vskip -0.05in
    \caption{The shared subspace of two networks trained on CIFAR-10 from different random initializations is spanned primarily by the eigenvectors corresponding to the largest eigenvalues. Each row represents a different network layer. Note that the average pooling layer has only 64 units. \textbf{Left}: Scaling of the eigenvectors $\mathbf{u}_X^i$ of the RSM $XX^\text{T}$ from network A by RSMs of networks A and B. Orange lines show $||XX^\text{T}\mathbf{u}_X^i||_2$, \ie~the eigenvalues. Purple dots show $||YY^\text{T}\mathbf{u}_X^i||_2$, the scaling of the eigenvectors of the RSM of network $A$ by the RSM of network $B$. \textbf{Right}: Cosine of the rotation by the RSM of network B, $(\mathbf{u}_X^i)^\text{T}YY^\text{T}\mathbf{u}_X^i/||YY^\text{T}\mathbf{u}_X^i||_2$.}
    \label{fig:eigs}
    \vskip -0.1in
\end{figure}

Equation \ref{eq:kta_eigs} suggests a way to further elucidating what CKA is measuring, based on the action of one representational similarity matrix (RSM) $YY^\text{T}$ applied to the eigenvectors $\mathbf{u}_X^i$ of the other RSM $XX^\text{T}$. By definition, $XX^\text{T}\mathbf{u}_X^i$ points in the same direction as $\mathbf{u}_X^i$, and its norm $||XX^\text{T}\mathbf{u}_X^i||_2$ is the corresponding eigenvalue. The degree of scaling and rotation by $YY^\text{T}$ thus indicates how similar the action of $YY^\text{T}$ is to $XX^\text{T}$, for each eigenvector of $XX^\text{T}$. For visualization purposes, this approach is somewhat less useful than the CKA summary statistic, since it does not collapse the similarity to a single number, but it provides a more complete picture of what CKA measures. Figure \ref{fig:eigs} shows that, for large eigenvectors, $XX^\text{T}$ and $YY^\text{T}$ have similar actions, but the rank of the subspace where this holds is substantially lower than the dimensionality of the activations. In the penultimate (global average pooling) layer, the dimensionality of the shared subspace is approximately 10, which is the number of classes in the CIFAR-10 dataset.

\section{Conclusion and Future Work}
Measuring similarity between the representations learned by neural networks is an ill-defined problem, since it is not entirely clear what aspects of the representation a similarity index should focus on. Previous work has suggested that there is little similarity between intermediate layers of neural networks trained from different random initializations \cite{svcca,wang2018}. We propose CKA as a method for comparing representations of neural networks, and show that it consistently identifies correspondences between layers, not only in the same network trained from different initializations, but across entirely different architectures, whereas other methods do not. 
We also provide a unified framework for understanding the space of similarity indexes, as well as an empirical framework for evaluation.

We show that CKA captures intuitive notions of similarity, \ie~that neural networks trained from different initializations should be similar to each other. However, it remains an open question whether there exist kernels beyond the linear and RBF kernels that would be better for analyzing neural network representations. Moreover, there are other potential choices of weighting in Equation \ref{eq:kta_eigs} that may be more appropriate in certain settings. We leave these questions as future work. 
Nevertheless, CKA seems to be much better than previous methods at finding correspondences between the learned representations in hidden layers of neural networks.

\section*{Acknowledgements}
We thank Gamaleldin Elsayed, Jaehoon Lee, Paul-Henri Mignot, Maithra Raghu, Samuel L. Smith, Alex Williams, and Michael Wu for comments on the manuscript, Rishabh Agarwal for ideas, and Aliza Elkin for support.

\bibliography{main}
\bibliographystyle{icml2019}

\cleardoublepage
\appendix
\counterwithin{figure}{section}
\counterwithin{table}{section}
\FloatBarrier\onecolumn

\section{Proof of Theorem 1}
\label{theorem_1_proof}
\begin{theorem*}
Let $X$ and $Y$ be $n \times p$ matrices. Suppose $s$ is invariant to invertible linear transformation in the first argument, \ie~$s(X, Z) = s(XA, Z)$ for arbitrary $Z$ and any $A$ with $\text{rank}(A) = p$. If $\text{rank}(X) = \text{rank}(Y) = n$, then $s(X, Z) = s(Y, Z)$.

\begin{proof}
Let
\begin{align*}
X' &= \begin{bmatrix}
X\\
K_X
\end{bmatrix} & Y' &= \begin{bmatrix}
Y\\
K_Y 
\end{bmatrix} ,
\end{align*}
where $K_X$ is a basis for the null space of the rows of $X$ and $K_Y$ is a basis for the null space of the rows of $Y$. Then let $A = X'^{-1}Y'$.
$$\begin{bmatrix}
X\\
K_X
\end{bmatrix} A = \begin{bmatrix}
Y\\
K_Y 
\end{bmatrix} \implies XA = Y .$$
Because $X'$ and $Y'$ have rank $p$ by construction, $A$ also has rank $p$. Thus, $s(X, Z) = s(XA, Z) = s(Y, Z)$.
\end{proof}
\end{theorem*}

\section{Orthogonalization and Invariance to Invertible Linear Transformation}
\label{orthogonality_invariance_theorem}

Here we show that any similarity index that is invariant to orthogonal transformation can be made invariant to invertible linear transformation by orthogonalizing the columns of the input.

\begin{proposition}
Let $X$ be an $n \times p$ matrix of full column rank and let $A$ be an invertible $p \times p$ matrix. Let $X = Q_XR_X$ and $XA = Q_{XA} R_{XA}$, where $Q_X^{T}Q_X = Q_{XA}^{T}Q_{XA} = I$ and $R_X$ and $R_{XA}$ are invertible. If $s(\cdot,\cdot)$ is invariant to orthogonal transformation,
then $s(Q_X, Y) = s(Q_{XA}, Y)$.

\begin{proof}
Let $B = R_{X} A R_{XA}^{-1}$. Then $Q_X B = Q_{XA}$, and B is an orthogonal transformation:
$$B^\text{T} B =  B^\text{T} Q_X^\text{T}Q_X B = Q_{XA}^\text{T} Q_{XA} = I .$$
Thus $s(Q_{X}, Y) = s(Q_{X}B, Y) = s(Q_{XA}, Y)$.
\end{proof}
\end{proposition}

\section{CCA and Linear Regression}
\label{cca_and_linear_regression}
\subsection{Linear Regression}
Consider the linear regression fit of the columns of an $n \times m$ matrix $C$ with an $n \times p$ matrix $A$:
\begin{align*}
    \hat B = \argmin_B ||C - A B||_\text{F}^2 = (A^\text{T}A)^{-1} A^\text{T} C .
\end{align*}
Let $A = Q R$, the thin QR decomposition of A. Then the fitted values are given by:
\begin{align*}
    \hat C &= A \hat B\\
    &= A (A^\text{T}A)^{-1}A^\text{T} C\\
    &= Q R (R^\text{T}Q^\text{T}Q R)^{-1}R^\text{T}Q^\text{T} C\\ 
    &= Q R R^{-1} (R^\text{T})^{-1} R^\text{T}Q^\text{T} C\\
    &= Q Q^\text{T} C .
\end{align*}
The residuals $E = C - \hat C$ are orthogonal to the fitted values, \ie{}
\begin{align*}
  E^\text{T} \hat C &= (C - Q Q^\text{T} C)^\text{T} Q Q^\text{T} C\\
  &= C^\text{T} Q Q^\text{T} C - C^\text{T} Q Q^\text{T} C = 0 .
\end{align*}
Thus:
\begin{align}
    ||E||_\text{F}^2 &= \text{tr}(E^\text{T} E)\nonumber\\
    &= \text{tr}(E^\text{T} C - E^\text{T} \hat C)\nonumber\\
    &= \text{tr}((C - \hat C)^\text{T} C)\nonumber\\
    &= \text{tr}(C^\text{T} C) - \text{tr}(C^\text{T} Q Q^\text{T} C)\nonumber\\
    &= ||C||_\text{F}^2 - ||Q^\text{T} C||_\text{F}^2 .
\end{align}
Assuming that $C$ was centered by subtracting its column means prior to the linear regression fit, the total fraction of variance explained by the fit is:
\begin{align}
R^2 &= 1 - \frac{||E||_\text{F}^2}{||C||_\text{F}^2} = 1 - \frac{||C||_\text{F}^2 - ||Q^\text{T} C||_\text{F}^2}{||C||_\text{F}^2} = \frac{||Q^\text{T} C||_\text{F}^2}{||C||_\text{F}^2} \label{r2} .
\end{align}
Although we have assumed that $Q$ is obtained from QR decomposition, any orthonormal basis with the same span will suffice, because orthogonal transformations do not change the Frobenius norm.

\subsection{CCA}
Let $X$ be an $n \times p_1$ matrix and $Y$ be an $n \times p_2$ matrix, and let $p = \text{min}(p_1, p_2)$. Given the thin QR decompositions of $X$ and $Y$, $X = Q_X R_X$, $Y = Q_Y R_Y$ such that $Q_X^\text{T} Q_X = I$, $Q_Y^\text{T} Q_Y = I$, the canonical correlations $\rho_i$ are the singular values of $A = Q_X^\text{T} Q_Y$ \cite{bjorck1973numerical,press2011canonical} and thus the square roots of the eigenvalues of $A^\text{T} A$. The squared canonical correlations $\rho_i^2$ are the eigenvalues of $A^\text{T} A = Q_Y^\text{T} Q_X Q_X^\text{T} Q_Y$. Their sum is $\sum_{i=1}^{p} \rho_i^2 = \text{tr}(A^\text{T} A) = ||Q_Y^\text{T} Q_X||_\text{F}^2$.

Now consider the linear regression fit of the columns of $Q_X$ with $Y$. Assume that $Q_X$ has zero mean. Substituting $Q_Y$ for $Q$ and $Q_X$ for $C$ in Equation \ref{r2}, and noting that $||Q_X||_\text{F}^2 = p_1$:
\begin{align}
R^2 = \frac{||Q_Y^\text{T} Q_X||_\text{F}^2}{p_1} = \frac{\sum_{i=1}^{p} \rho_i^2}{p_1} .
\end{align}

\subsection{Projection-Weighted CCA}
\label{pwcca}

Let $X$ be an $n \times p_1$ matrix and $Y$ be an $n \times p_2$ matrix, with $p_1 \leq p_2$. \citet{morcos2018insights} proposed to compute projection-weighted canonical correlation as:
\begin{align*}
\bar \rho_\text{PW} &= \frac{\sum_{i=1}^c \alpha_i \rho_i}{\sum_{i=1} \alpha_i} & 
\alpha_i &=\sum_{j} |\langle \mathbf{h}_i, \mathbf{x}_j \rangle| ,
\end{align*} where the $\mathbf{x}_j$ are the columns of $X$, and the $\mathbf{h}_i$ are the canonical variables formed by projecting $X$ to the canonical coordinate frame. Below, we show that if we modify $\bar \rho_\text{PW}$ by squaring the dot products and $\rho_i$, we recover linear regression.:
\begin{align*}
R_\text{MPW}^2 &= \frac{\sum_{i=1}^c \alpha_i' \rho_i^2}{\sum_{i=1} \alpha_i'} = R_\text{LR}^2 & 
\alpha_i' &=\sum_{j} \langle \mathbf{h}_i, \mathbf{x}_j \rangle^2 .
\end{align*}
Our derivation begins by forming the SVD $Q_X^\text{T} Q_Y = U \Sigma V^\text{T}$. $\Sigma$ is a diagonal matrix of the canonical correlations $\rho_i$, and the matrix of canonical variables $H = Q_X U$. Then $R^2_\text{MPW}$ is:
\begin{align}
    R_\text{MPW}^2 &= \frac{||X^\text{T} H \Sigma||_\text{F}^2}{||X^\text{T} H||_\text{F}^2}\\
    &= \frac{\text{tr}((X^\text{T} H \Sigma)^\text{T}(X^\text{T} H \Sigma))}{\text{tr}((X^\text{T} H)^\text{T}(X^\text{T} H))}\nonumber\\
    &= \frac{\text{tr}(\Sigma H^\text{T} X X^\text{T} H \Sigma)}{\text{tr}(H^\text{T} X  X^\text{T} H)}\nonumber\\
    &= \frac{\text{tr}(X^\text{T} H \Sigma^2 H^\text{T} X)}{\text{tr}( X^\text{T} H H^\text{T}  X)}\nonumber\\
    &= \frac{\text{tr}(R_X^\text{T} Q_X^\text{T} H \Sigma^2 H^\text{T} Q_X R_X)}{\text{tr}(R_X^\text{T} Q_X^\text{T} Q_X U U^\text{T} Q_X^\text{T} Q_X R_X)} . \nonumber
\end{align}
Because we assume $p_1 \leq p_2$, $U$ is a square orthogonal matrix and $UU^\text{T} = I$. Further noting that $Q_X^\text{T} H = U$ and $U \Sigma = Q_X^\text{T} Q_Y V$:
\begin{align}
     R_\text{MPW}^2 &= \frac{\text{tr}(R_X^\text{T} U \Sigma^2 U^\text{T} R_X)}{\text{tr}(R_X^\text{T} Q_X^\text{T} Q_X R_X)}\nonumber\\
    &= \frac{\text{tr}(R_X^\text{T} Q_X^\text{T} Q_Y V \Sigma U^\text{T} R_X)}{\text{tr}(X^\text{T} X)}\nonumber\\
    &= \frac{\text{tr}(X^\text{T} Q_Y Q_Y^\text{T} Q_X R_X)}{\text{tr}(X^\text{T} X)}\nonumber\\
    &= \frac{\text{tr}(X^\text{T} Q_Y Q_Y^\text{T} X)}{\text{tr}(X^\text{T} X)}\nonumber\\
    &= \frac{||Q_Y^\text{T} X||_\text{F}^2}{||X||_\text{F}^2} . \nonumber
\end{align}
Substituting $Q_Y$ for $Q$ and $X$ for $C$ in Equation \ref{r2}:
$$R_\text{LR}^2 = \frac{||Q_Y^\text{T} X||_\text{F}^2}{||X||_\text{F}^2} = R_\text{MPW}^2 .$$

\section{Notes on Other Methods}

\subsection{Canonical Ridge}
\label{appendix:regularized_cca}
Beyond CCA, we could also consider the ``canonical ridge" regularized CCA objective \cite{vinod1976canonical}:
\begin{equation}
\begin{aligned}
\label{eq:regularized_cca}
\sigma_i = \max_{\mathbf{w}_X^i, \mathbf{w}_Y^i}&  \frac{(X\mathbf{w}_X^i)^\text{T}(Y\mathbf{w}_Y^i)}{\sqrt{||X\mathbf{w}_X^i||^2 + \kappa_X ||\mathbf{w}_X^i||_2^2}\sqrt{||Y\mathbf{w}_Y^i||^2 + \kappa_Y ||\mathbf{w}_Y^i||^2}}\\
\mathrm{subject~to} &\ ~~ \forall_{j< i}~~ (\textbf{w}_X^i)^\text{T} (X^\text{T} X + \kappa I) \textbf{w}_X^j = 0\\
&\ ~~ \forall_{j< i}~~ (\textbf{w}_Y^i)^\text{T} (Y^\text{T} Y + \kappa I) \textbf{w}_Y^j = 0 .
\end{aligned}
\end{equation}

Given the singular value decompositions $X = U_X\Sigma_XV_X^\text{T}$ and $Y = U_Y\Sigma_YV_Y^\text{T}$, one can form ``partially orthogonalized" bases $\tilde Q_X = U_X\Sigma_X(\Sigma_X^2 + \kappa_X I)^{-1/2}$ and $\tilde Q_Y = U_Y\Sigma_Y(\Sigma_Y^2 + \kappa_Y I)^{-1/2}$. Given the singular value decomposition of their product $\tilde U \tilde \Sigma \tilde V^\text{T} = \tilde Q_X^\text{T} \tilde Q_Y$, the canonical weights are given by $W_X = V_X(\Sigma_X^2 + \kappa_X I)^{-1/2}\tilde U$ and $W_Y = V_Y(\Sigma_Y^2 + \kappa_Y I)^{-1/2}\tilde V$, as previously shown by \citet{mroueh2015asymmetrically}. As in the unregularized case (Equation \ref{eq:cca_eigs}), there is a convenient expression for the sum of the squared singular values $\sum \tilde \sigma_i^2$ in terms of the eigenvalues and eigenvectors of $XX^\text{T}$ and $YY^\text{T}$. Let the $i^\text{th}$ left-singular vector of $X$ (eigenvector of $XX^\text{T}$) be indexed as $\textbf{u}_X^\text{i}$ and let the $i^\text{th}$ eigenvalue of $XX^\text{T}$ (squared singular value of $X$) be indexed as $\lambda_X^i$, and similarly let the left-singular vectors of $YY^\text{T}$ be indexed as $\textbf{u}_Y^\text{i}$ and the eigenvalues as $\lambda_Y^i$. Then:
\begin{align}
\sum_{i=1}^{p_1} \tilde \sigma_i^2 &= ||\tilde Q_Y^\text{T} \tilde Q_X||_\text{F}^2\\
&= ||(\Sigma_Y^2 + \kappa_Y I)^{-1/2}\Sigma_Y U_Y^\text{T} U_X\Sigma_X(\Sigma_X^2 + \kappa_X I)^{-1/2}||_\text{F}^2\\
&= \sum_{i=1}^{p_1} \sum_{j=1}^{p_2} \frac{\lambda_X^i \lambda_Y^j }{(\lambda_X^i + \kappa_X)(\lambda_Y^j + \kappa_Y)} \langle\textbf{u}_X^i, \textbf{u}_Y^j\rangle^2 .
\label{eq:reg_eigs}
\end{align}

Unlike in the unregularized case, the singular values $\sigma_i$ do not measure the correlation between the canonical variables. Instead, they become arbitrarily small as $\kappa_X$ or $\kappa_Y$ increase. Thus, we need to normalize the statistic to remove the dependency on the regularization parameters.

Applying von Neumann's trace inequality yields a bound:
\begin{align}
    \sum_{i=1}^{p_1} \tilde \sigma_i^2 &= \text{tr}(\tilde Q_Y \tilde Q_Y^\text{T} \tilde Q_X \tilde Q_X^\text{T})\\
    &= \text{tr}((U_Y\Sigma_Y^2(\Sigma_Y^2 + \kappa_Y I)^{-1} U_Y^\text{T})(U_X\Sigma_X^2(\Sigma_X^2 + \kappa_X I)^{-1}U_X^\text{T}))\\
    &\leq \sum_{i=1}^{p_1} \frac{\lambda_X^i \lambda_Y^i}{(\lambda_X^i + \kappa_X)(\lambda_Y^i + \kappa_Y)} . \label{eq:von_neumann_norm1}
\end{align}
Applying the Cauchy-Schwarz inequality to \eqref{eq:von_neumann_norm1} yields the alternative bounds:
\begin{align}
    \sum_{i=1}^{p_1} \tilde \sigma_i^2 &\leq \sqrt{\sum_{i=1}^{p_1} \left(\frac{\lambda_X^i}{\lambda_X^i + \kappa_X}\right)^2}\sqrt{\sum_{i=1}^{p_1} \left(\frac{\lambda_Y^i}{\lambda_Y^i + \kappa_Y}\right)^2}\label{eq:von_neumann_norm2}\\
    &\leq \sqrt{\sum_{i=1}^{p_1} \left(\frac{\lambda_X^i}{\lambda_X^i + \kappa_X}\right)^2}\sqrt{\sum_{i=1}^{p_2} \left(\frac{\lambda_Y^i}{\lambda_Y^i + \kappa_Y}\right)^2} . \label{eq:von_neumann_norm3}
\end{align}
A normalized form of \eqref{eq:reg_eigs} could be produced by dividing by any of \eqref{eq:von_neumann_norm1}, \eqref{eq:von_neumann_norm2}, or \eqref{eq:von_neumann_norm3}.

If $\kappa_X = \kappa_Y = 0$, then \eqref{eq:von_neumann_norm1} and \eqref{eq:von_neumann_norm2} are equal to $p_1$. In this case, \eqref{eq:reg_eigs} is simply the sum of the squared canonical correlations, so normalizing by either of these bounds recovers $R^2_\text{CCA}$.

If $\kappa_Y = 0$, then as $\kappa_X \to \infty$, normalizing by the bound from \eqref{eq:von_neumann_norm1} recovers $R^2$:
\begin{align}
    & \lim_{\kappa_X \to \infty} \frac{\sum_{i=1}^{p_1} \sum_{j=1}^{p_2} \frac{\lambda_X^i \lambda_Y^j }{(\lambda_X^i + \kappa_X)(\lambda_Y^j + 0)} \langle\textbf{u}_X^i, \textbf{u}_Y^j\rangle^2}{\sum_{i=1}^{p_1} \frac{\lambda_X^i \lambda_Y^i}{(\lambda_X^i + \kappa_X)(\lambda_Y^i + 0)}}\\
    =& \lim_{\kappa_X \to \infty} \frac{\sum_{i=1}^{p_1} \sum_{j=1}^{p_2} \frac{\lambda_X^i}{\left(\frac{\lambda_X^i}{\kappa_X} + 1\right)} \langle\textbf{u}_X^i, \textbf{u}_Y^j\rangle^2}{\sum_{i=1}^{p_1} \frac{\lambda_X^i}{\left(\frac{\lambda_X^i}{\kappa_X} + 1\right)}}\\
    =& \frac{\sum_{i=1}^{p_1} \sum_{j=1}^{p_2} \lambda_X^i \langle\textbf{u}_X^i, \textbf{u}_Y^j\rangle^2}{\sum_{i=1}^{p_1} \lambda_X^i}\\
    =& \frac{||U_Y^\text{T}U_X\Sigma_X||_\text{F}^2}{||X||_\text{F}^2} = \frac{||Q_Y^\text{T}X||_\text{F}^2}{||X||_\text{F}^2} = R^2_\text{LR} .
\end{align}

The bound from \eqref{eq:von_neumann_norm3} differs from the bounds in \eqref{eq:von_neumann_norm1} and \eqref{eq:von_neumann_norm2} because it is multiplicatively separable in $X$ and $Y$. Normalizing by this bound leads to $\text{CKA}(\tilde Q_X \tilde Q_X^\text{T}, \tilde Q_Y \tilde Q_Y^\text{T})$:
\begin{align}
& \frac{\sum_{i=1}^{p_1} \sum_{j=1}^{p_2} \frac{\lambda_X^i \lambda_Y^j }{(\lambda_X^i + \kappa_X)(\lambda_Y^j + \kappa_Y)} \langle\textbf{u}_X^i, \textbf{u}_Y^j\rangle^2}{\sqrt{\sum_{i=1}^{p_1} \left(\frac{\lambda_X^i}{\lambda_X^i + \kappa_X}\right)^2}\sqrt{\sum_{i=1}^{p_2} \left(\frac{\lambda_Y^i}{\lambda_Y^i + \kappa_Y}\right)^2}}\\
=&\ \frac{||\tilde Q_Y^\text{T}\tilde Q_X||_\text{F}^2}{||\tilde Q_X^\text{T}\tilde Q_X||_\text{F}||\tilde Q_Y^\text{T}\tilde Q_Y||_\text{F}} = \text{CKA}(\tilde Q_X \tilde Q_X^\text{T}, \tilde Q_Y \tilde Q_Y^\text{T}) .
\end{align}

Moreover, setting $\kappa_X = \kappa_Y = \kappa$ and taking the limit as $\kappa \to \infty$, the normalization from \eqref{eq:von_neumann_norm3} leads to $\text{CKA}(XX^\text{T}, YY^\text{T})$:
\begin{align}
    & \lim_{\kappa \to \infty} \frac{\sum_{i=1}^{p_1} \sum_{j=1}^{p_2} \frac{\lambda_X^i \lambda_Y^j }{(\lambda_X^i + \kappa)(\lambda_Y^j + \kappa)} \langle\textbf{u}_X^i, \textbf{u}_Y^j\rangle^2}{\sqrt{\sum_{i=1}^{p_1} \left(\frac{\lambda_X^i}{\lambda_X^i + \kappa}\right)^2}\sqrt{\sum_{i=1}^{p_2} \left(\frac{\lambda_Y^i}{\lambda_Y^i + \kappa}\right)^2}}\\
    =& \lim_{\kappa \to \infty} \frac{\sum_{i=1}^{p_1} \sum_{j=1}^{p_2} \frac{\lambda_X^i \lambda_Y^j }{\left(\frac{\lambda_X^i}{\kappa} + 1\right)\left(\frac{\lambda_Y^j}{\kappa} + 1\right)} \langle\textbf{u}_X^i, \textbf{u}_Y^j\rangle^2}{\sqrt{\sum_{i=1}^{p_1} \left(\frac{\lambda_X^i}{\frac{\lambda_X^i}{\kappa} + 1}\right)^2}\sqrt{\sum_{i=1}^{p_2} \left(\frac{\lambda_Y^i}{\frac{\lambda_Y^i}{\kappa} + 1}\right)^2}}\\
    =&\ \frac{\sum_{i=1}^{p_1} \sum_{j=1}^{p_2} \lambda_X^i \lambda_Y^j \langle\textbf{u}_X^i, \textbf{u}_Y^j\rangle^2}{\sqrt{\sum_{i=1}^{p_1} \left(\lambda_X^i\right)^2}\sqrt{\sum_{i=1}^{p_2} \left(\lambda_Y^i\right)^2}}\\
    =&\ \text{CKA}(XX^\text{T}, YY^\text{T}) . \nonumber
\end{align}

Overall, the hyperparameters of the canonical ridge objective make it less useful for exploratory analysis. These hyperparameters could be selected by cross-validation, but this is computationally expensive, and the resulting estimator would be biased by sample size. Moreover, our goal is not to map representations of networks to a common space, but to measure the similarity between networks. Appropriately chosen regularization will improve out-of-sample performance of the mapping, but it makes the meaning of ``similarity" more ambiguous.

\subsection{The Orthogonal Procrustes Problem}
The orthogonal Procrustes problem consists of finding an orthogonal rotation in feature space that produces the smallest error:
\begin{align}
    \hat Q = \argmin_Q ||Y - X Q||_\text{F}^2~~\mathrm{subject~to}~~Q^\text{T}Q = I .
\end{align}
The objective can be written as:
\begin{align}
    ||Y - X Q||_\text{F}^2 &= \text{tr}((Y - XQ)^\text{T}(Y - XQ)) \nonumber\\
    &= \text{tr}(Y^\text{T}Y) - \text{tr}(Y^\text{T}XQ) - \text{tr}(Q^\text{T}X^\text{T}Y) + \text{tr}(Q^\text{T}X^\text{T}XQ) \nonumber\\
    &= ||Y||_\text{F}^2 + ||X||_\text{F}^2 - 2\text{tr}(Y^\text{T}XQ) .
\end{align}
Thus, an equivalent objective is:
\begin{align}
    \hat Q = \argmax_Q \text{tr}(Y^\text{T}XQ)~~\mathrm{subject~to}~~Q^\text{T}Q = I .
    \label{eq:procrustes_trace}
\end{align}
The solution is $\hat Q = U V^\text{T}$ where $U\Sigma V^\text{T} = X^\text{T} Y$, the singular value decomposition. At the maximum of \eqref{eq:procrustes_trace}:
\begin{align}
    \text{tr}(Y^\text{T}X\hat{Q}) = \text{tr}(V\Sigma U^\text{T} U V^\text{T}) = \text{tr}(\Sigma) = ||X^\text{T}Y||_* = ||Y^\text{T}X||_* ,
\end{align}
which is similar to what we call ``dot product-based similarity" (Equation~\ref{linear_ss}), but with the squared Frobenius norm of $Y^\text{T}X$ (the sum of the squared singular values) replaced by the nuclear norm (the sum of the singular values). The Frobenius norm of $Y^\text{T}X$ can be obtained as the solution to a similar optimization problem:
\begin{align}
    ||Y^\text{T}X||_\text{F} &= \max_W \text{tr}(Y^\text{T}XW)~~\mathrm{subject~to}~~ \text{tr}(W^\text{T}W) = 1 .
\end{align}

In the context of neural networks, \citet{smith2017offline} previously proposed using the solution to the orthogonal Procrustes problem to align word embeddings from different languages, and demonstrated that it outperformed CCA.

\section{Architecture Details}
\label{architecture_details}
All non-ResNet architectures are based on All-CNN-C \cite{springenberg2014striving}, but none are architecturally identical. The Plain-10 model is very similar, but we place the final linear layer after the average pooling layer and use batch normalization because these are common choices in modern architectures. We use these models because they train in minutes on modern hardware.

\begin{table}[h]
    \centering
    \begin{center}
    \begin{small}
    \begin{tabular}{p{5cm}}
        \toprule
        \multicolumn{1}{c}{Tiny-10}\\
        \midrule
        $3\times 3$ conv. 16-BN-ReLu $\times 2$\\
        $3\times 3$ conv. 32 stride 2-BN-ReLu\\
        $3\times 3$ conv. 32-BN-ReLu $\times 2$\\
        $3\times 3$ conv. 64 stride 2-BN-ReLu\\
        $3\times 3$ conv. 64 valid padding-BN-ReLu\\
        $1\times 1$ conv. 64-BN-ReLu\\
        Global average pooling\\
        Logits\\
        \bottomrule
    \end{tabular}
    \end{small}
    \end{center}
    \caption{The Tiny-10 architecture, used in Figures \ref{fig:comparison}, \ref{fig:eigs}, \ref{fig:other_thresholds}, and \ref{app:tiny_10_v_resnet}. The average Tiny-10 model achieved 89.4\% accuracy.}
    \label{tab:tiny_arch}
\end{table}

\begin{table}[h]
    \centering
    \vskip 0.15in
    \begin{center}
    \begin{small}
    \begin{tabular}{p{5cm}}
        \toprule
        \multicolumn{1}{c}{Plain-$(8n+2)$}\\
        \midrule
        $3\times 3$ conv. 96-BN-ReLu $\times (3n-1)$\\
        $3\times 3$ conv. 96 stride 2-BN-ReLu\\
        $3\times 3$ conv. 192-BN-ReLu $\times (3n-1)$\\
        $3\times 3$ conv. 192 stride 2-BN-ReLu\\
        $3\times 3$ conv. 192 BN-ReLu $\times (n-1)$\\
        $3\times 3$ conv. 192 valid padding-BN-ReLu\\
        $1\times 1$ conv. 192-BN-ReLu $\times n$\\
        Global average pooling\\
        Logits\\
        \bottomrule
    \end{tabular}
    \end{small}
    \end{center}
    \caption{The Plain-$(8n+2)$ architecture, used in Figures \ref{fig:different_depths}, \ref{fig:cross_arch}, \ref{fig:cross_dataset},  \ref{fig:init}, \ref{fig:init_other_arch}, \ref{fig:bn}, and  \ref{fig:within_class}. Mean accuracies: Plain-10, 93.9\%; Plain-18: 94.8\%; Plain-34: 93.7\%; Plain-66: 91.3\%}
    \label{tab:plain_arch}
\end{table}

\begin{table}[h]
    \centering
    \vskip 0.15in
    \begin{center}
    \begin{small}
    \begin{tabular}{p{5cm}}
        \toprule
        \multicolumn{1}{c}{Width-$n$}\\
        \midrule
        $3\times 3$ conv. $n$-BN-ReLu $\times 2$\\
        $3\times 3$ conv. $n$ stride 2-BN-ReLu\\
        $3\times 3$ conv. $n$-BN-ReLu $\times 2$\\
        $3\times 3$ conv. $n$ stride 2-BN-ReLu\\
        $3\times 3$ conv. $n$ valid padding-BN-ReLu\\
        $1\times 1$ conv. $n$-BN-ReLu\\
        Global average pooling\\
        Logits\\
        \bottomrule
    \end{tabular}
    \end{small}
    \end{center}
    \caption{The architectures used for width experiments in Figure~\ref{fig:width_experiments}.}
    \label{tab:width_arch}
\end{table}
\clearpage
\section{Additional Experiments}
\subsection{Sanity Check for Transformer Encoders}
\label{appendix:transformer}

\begin{figure}[h!]
    \begin{minipage}[c]{0.49\linewidth}
        \centering
        \includegraphics[width=3.25in]{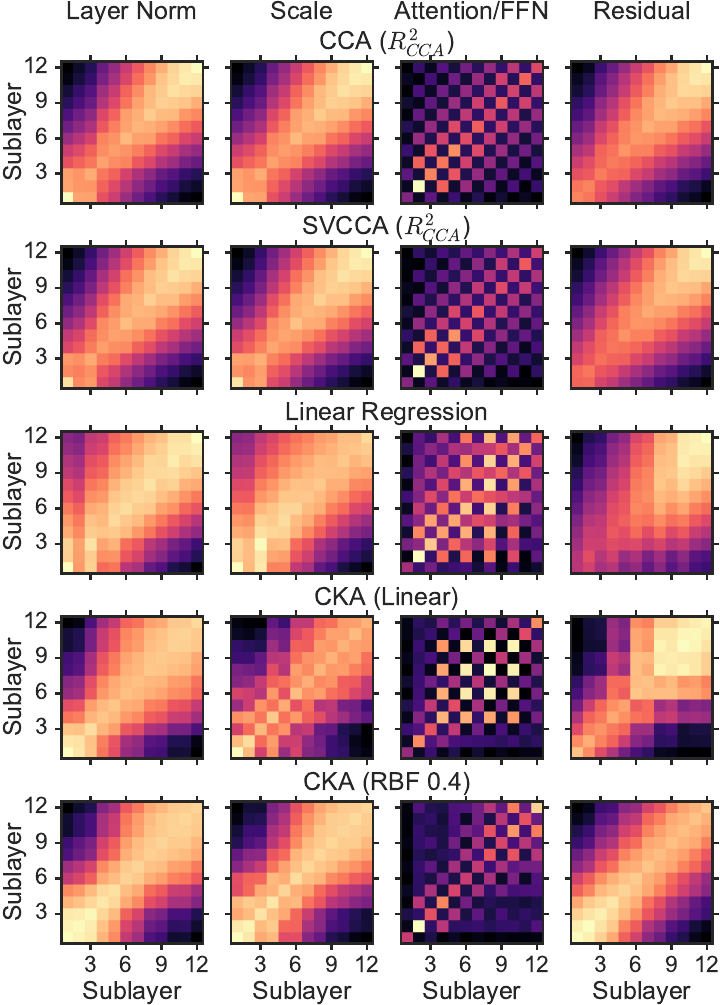}
        \caption{All similarity indices broadly reflect the structure of Transformer encoders. Similarity indexes are computed between the 12 sublayers of Transformer encoders, for each of the 4 possible places in each sublayer that representations may be taken (see Figure~\ref{fig:transformer_architecture}), averaged across 10 models trained from different random initializations.}
        \label{fig:transformer}
    \end{minipage}\hfill
    \begin{minipage}[c]{0.49\linewidth}
        \begin{minipage}[c]{\linewidth}
            \centering
                \includegraphics[width=1.42in]{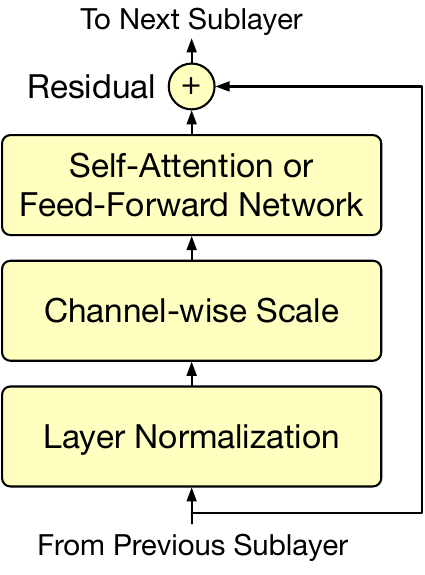}
            \caption{Architecture of a single sublayer of the Transformer encoder used for our experiments. The full encoder includes 12 sublayers, alternating between self-attention and feed-forward network sublayers.}
            \vskip 0.2in
            \label{fig:transformer_architecture}
        \end{minipage}
        \begin{minipage}[c]{\linewidth}
            \centering
            \begin{small}
            {
            \setlength{\tabcolsep}{0.4em} %
            \begin{tabular}{lrrrr}
                \toprule
                Index & Layer Norm & Scale & Attn/FFN & Residual\\
                 \midrule
                CCA ($\bar \rho$) & 85.3 & 85.3 & \textbf{94.9} & 90.9\\
                CCA ($R^2_\text{CCA}$) & 87.8 & 87.8 & \textbf{95.3} & \textbf{95.2}\\
                SVCCA ($\bar \rho$) & 78.2 & 83.0 & 89.5 & 75.9\\
                SVCCA ($R^2_\text{CCA}$) & 85.4 & 86.9 & 90.8 & 84.7\\
                PWCCA & \textbf{88.5} & 88.9 & \textbf{96.1} & 87.0\\
                Linear Reg. & 78.1 & 83.7 & 76.0 & 36.9\\
                CKA (Linear) & 78.6 & \textbf{95.6} & 86.0 & 73.6\\
                CKA (RBF 0.2) & 76.5 & 73.1 & 70.5 & 76.2\\
                CKA (RBF 0.4) & \textbf{92.3} & \textbf{96.5} & 89.1 & \textbf{98.1}\\
                CKA (RBF 0.8) & 80.8 & \textbf{95.8} & \textbf{93.6} & 90.0\\
                \bottomrule
            \end{tabular}
            }
            \captionof{table}{Accuracy of identifying corresponding sublayers based maximum similarity, for 10 architecturally identical 12-sublayer Transformer encoders at the 4 locations in each sublayer after which the representation may be taken (see Figure~\ref{fig:transformer_architecture}). Results not significantly different from the best result are bold-faced ($p < 0.05$, jackknife z-test).}
            \label{tab:transformer_eval}
            \end{small}
        \end{minipage}
    \end{minipage}
\end{figure}

When applied to Transformer encoders, all similarity indexes we investigated passed the sanity check described in Section~\ref{sec:sanity_check}. We trained Transformer models using the \verb!tensor2tensor! library \cite{vaswani2018tensor2tensor} on the English to German translation task from the WMT18 dataset \cite{bojar2018wmt} (Europarl v7, Common Crawl, and News Commentary v13 corpora) and computed representations of each of the 75,804 tokens from the 3,000 sentence \verb!newstest2013! development set, ignoring end of sentence tokens. In Figure~\ref{fig:transformer}, we show similarity between the 12 sublayers of the encoders of 10 Transformer models (45 pairs) trained from different random initializations. Each Transformer sublayer contains four operations, shown in Figure~\ref{fig:transformer_architecture}; results vary based which operation the representation is taken after. Table~\ref{tab:transformer_eval} shows the accuracy with which we identify corresponding layers between network pairs by maximal similarity.

The Transformer architecture alternates between self-attention and feed-forward network (FFN) sublayers. The checkerboard pattern in representational similarity after the self-attention/feed-forward network operation in Figure~\ref{fig:transformer} indicates that representations of attention sublayers are more similar to other attention sublayers than to FFN sublayers, and similarly, representations of FFN sublayers are more similar to other FFN than to feed-forward network layers. CKA reveals a checkerboard pattern for activations after the channel-wise scale operation (before the attention/FFN operation) that other methods do not. Because CCA is invariant to non-isotropic scaling, CCA similarities before and after channel-wise scaling are identical. Thus, CCA cannot capture this structure, even though the structure is consistent across networks.

\FloatBarrier
\subsection{SVCCA at Alternative Thresholds}
\label{app:other_thresholds}
\begin{figure}[h!]
    \centering
    \includegraphics[width=5.5in]{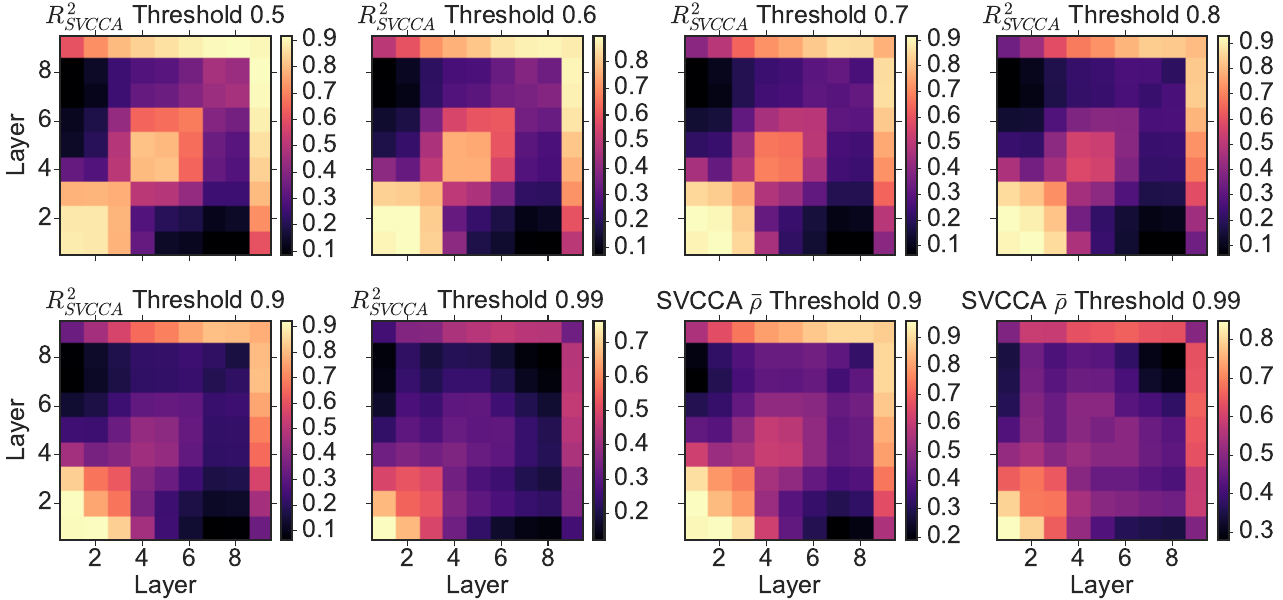}
    \vskip -0.05in
    \caption{Same as Figure \ref{fig:comparison} row 2, but for more SVCCA thresholds than the 0.99 threshold suggested by \citet{svcca}. No threshold reveals the structure of the network.}
    \label{fig:other_thresholds}
\end{figure}

\FloatBarrier
\subsection{CKA at Initialization}
\label{appendix:cka_init}
\begin{figure}[h!]
    \begin{minipage}[c]{\linewidth}
    \begin{minipage}[c]{0.485\linewidth}
    \centering
    \includegraphics[width=3.25in]{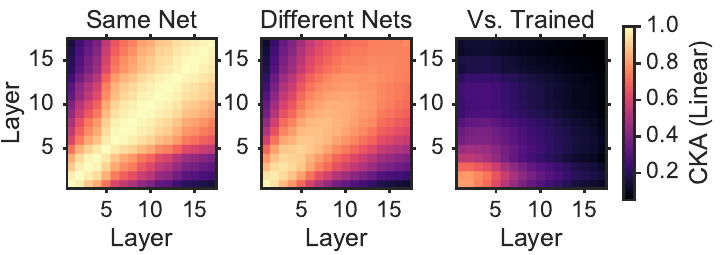}
    \vskip -0.05in
    \caption{Similarity of the Plain-18 network at initialization. \textbf{Left}: Similarity between layers of the same network. \textbf{Middle}: Similarity between untrained networks with different initializations. \textbf{Right}: Similarity between untrained and trained networks.}
    \label{fig:init}
    \end{minipage}\hfill
    \begin{minipage}[c]{0.485\linewidth}
    \centering
    \includegraphics[width=3.25in]{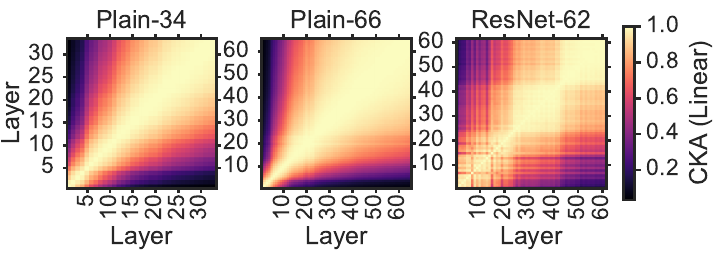}
    \vskip -0.05in
    \caption{Similarity between layers at initialization for deeper architectures.}
    \label{fig:init_other_arch}
    \end{minipage}
    \end{minipage}
\end{figure}

\FloatBarrier
\subsection{Additional CKA Results}
\begin{figure}[h!]
    \begin{minipage}[c]{\linewidth}
    \begin{minipage}[c]{0.485\linewidth}
    \centering
    \includegraphics[width=3.25in]{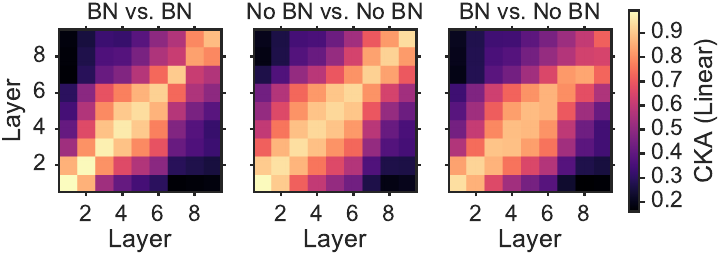}
    \vskip -0.05in
    \caption{Networks with and without batch normalization trained from different random initializations learn similar representations according to CKA. The largest difference between networks is at the last convolutional layer. Optimal hyperparameters were separately selected for the batch normalized network (93.9\% average accuracy) and the network without batch normalization (91.5\% average accuracy).}
    \label{fig:bn}
    \end{minipage}\hfill
    \begin{minipage}[c]{0.485\linewidth}
    \centering
    \includegraphics[width=3.25in]{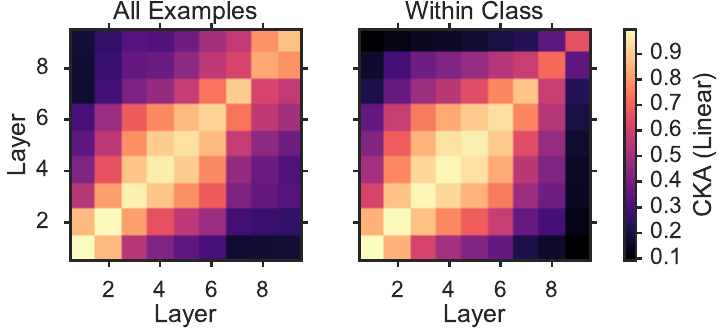}
    \vskip -0.1in
    \caption{Within-class CKA is similar to CKA based on all examples. To measure within-class CKA, we computed CKA separately for examples belonging to each CIFAR-10 class based on representations from Plain-10 networks, and averaged the resulting CKA values across classes.}
    \label{fig:within_class}
    \end{minipage}
    \end{minipage}
\end{figure}

\FloatBarrier
\subsection{Similarity Between Different Architectures with Other Indexes}
\label{app:tiny_10_v_resnet}
\begin{figure}[h!]
    \centering
    \includegraphics[width=6.75in]{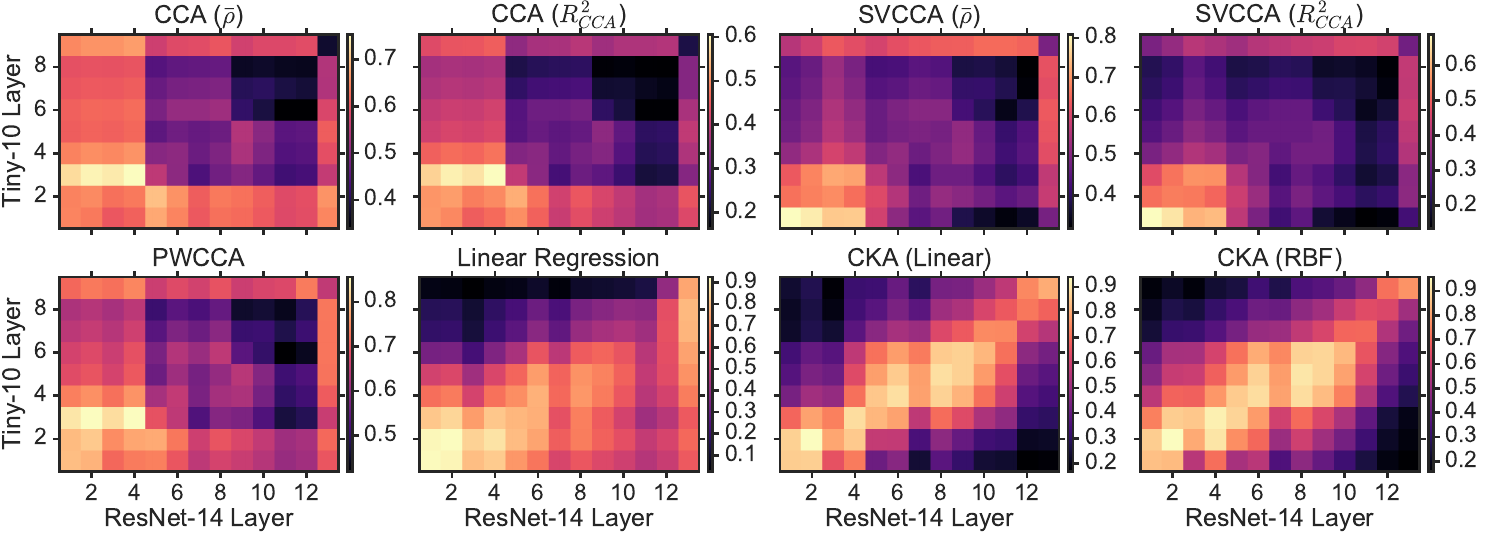}
    \vskip -0.05in
    \caption{Similarity between layers of different architectures (Tiny-10 and ResNet-14) for all methods investigated. Only CKA reveals meaningful correspondence. SVCCA results resemble Figure 7 of \citet{svcca}. In order to achieve better performance for CCA-based techniques, which are sensitive to the number of examples used to compute similarity, all plots show similarity on the CIFAR-10 training set.}
    \label{fig:tiny_10_v_resnet}
\end{figure}

\end{document}